# Exploring Body Texture from mmW Images for Person Recognition

Ester Gonzalez-Sosa, Ruben Vera-Rodriguez *Member, IEEE*, Julian Fierrez *Member, IEEE*,
Fernando Alonso-Fernandez, and Vishal M. Patel, *Senior Member, IEEE*

**Abstract**—Imaging using millimeter waves (mmWs) has many advantages including the ability to penetrate obscurants such as clothes and polymers. After having explored shape information retrieved from mmW images for person recognition, in this work we aim to gain some insight about the potential of using mmW texture information for the same task, considering not only the mmW face, but also mmW torso and mmW wholebody. We report experimental results using the mmW TNO database consisting of 50 individuals based on both hand-crafted and learned features from Alexnet and VGG-face pretrained Convolutional Neural Networks (CNN) models. First, we analyze the individual performance of three mmW body parts, concluding that: $i$) mmW torso region is more discriminative than mmW face and the whole body, $ii$) CNN features produce better results compared to hand-crafted features on mmW faces and the entire body, and $iii$) hand-crafted features slightly outperform CNN features on mmW torso. In the second part of this work, we analyze different multi-algorithmic and multi-modal techniques, including a novel CNN-based fusion technique, improving verification results to 2% EER and identification rank-1 results up to 99%. Comparative analyses with mmW body shape information and face recognition in the visible and NIR spectral bands are also reported.

## 1 INTRODUCTION

**M**ILLIMETER waves (mmW) are high-frequency electromagnetic waves usually defined to be in the range of 30 – 300 GHz with corresponding wavelengths between 10 to 1 mm. Since radiation at these frequencies is non-ionizing, it is considered to be safe for human exposure. Imaging using mmW has gained the interest of the security community [1], [2], [3], mainly due to its low intrusiveness and the ability to pass through clothing and other atmospheric obscurants such as cloud cover, fog, smoke, rain and dust storms.

Within the mmW spectrum range, there are two types of acquisition systems: active and passive. Active mmW scanners are those that emit mmW radiation to the scene and then collect the reflected radiation to form the image. In contrast, passive mmW scanners collect natural mmW radiation that has been emitted and reflected from the scene to form the image. Since passive mmW scanners collect natural mmW radiation, the ambient temperature affects the quality of image being formed. In particular, the images acquired by passive mmW scanners indoors are usually of low contrast than those acquired outdoors. In comparison, since the active mmW scanners rely on the transmitted and reflected radiation, the quality of the images acquired in different ambient temperature conditions are very similar. In general, higher quality images are obtained with active scanners than passive scanners. The predominant application of mmW images in the literature has been Concealed Weapon Detection (CWD), contraband or imaging under adverse weather conditions. In particular, the majority of international airports currently use active mmW scanners for detecting concealed objects. Exploration of mmW images for other purposes such as person recognition has been scarcely addressed in the literature. The privacy concerns of mmW images and the high cost of data acquisition are the two main obstacles that have prevented researchers from discovering new applications of mmW imaging.

*Ester Gonzalez-Sosa, Ruben Vera-Rodriguez and Julian Fierrez are with the Biometrics and Data Pattern Analytics Lab (BiDA) - ATVS, Escuela Politecnica Superior, Francisco Tomas y Valiente 11, Universidad Autonoma de Madrid, 28049 Madrid. e-mail: {ester.gonzalezs, ruben.vera, julian.fierrez}@uam.es*
*Vishal M. Patel is with the Department of Electrical and Computer Engineering, Rutgers University, Piscataway, NJ, 08854 USA e-mail: vishal.m.patel@rutgers.edu*
*Manuscript received ...*

Since millimeter waves can penetrate through clothing, in mmW images, we are able to see things that can not be seen in a visible image, such as the torso information. As a result, information collected in a mmW image may be useful for person recognition. This is specially important in scenarios where some kind of disguise or other presentation attacks can occur [4]. In such scenarios, standard biometric recognition may not be feasible [5], and working with mmW images may provide additional security against potential impostors. Motivated by the interest of the security community in these frequencies and the promising results obtained in the past few works for person recognition using mmW images [6], [7], we propose to use mmW images acquired from the screening scanners simultaneously for both detecting hidden objects and performing person recognition. We are aware that mmWs may not be the most convenient range of the spectrum for all the wide variety of person recognition applications, but we reckon it is worth exploiting the availability of these mmW scanners in security applications for further applications such as person recognition.

To this end, Alefs *et al.* [6] developed one of the first reported efforts for person recognition using real mmW passive images acquired in outdoors scenarios. They exploited the texture information contained in the torso region of the image through multilinear eigenspaces techniques. However, the experimental protocol carried out in that work was really optimistic, being far away from a realistic verification scenario in which a traveler would enter in a mmW scanner to verify his identity in comparison to a previously enrolled sample. On the other hand, the works by Moreno-Moreno *et al.* [8] and by Gonzalez-Sosa *et al.* [9] proposed and analyzed a biometric person recognition system using shape information, based on the idea that shape information retrieved from mmW images may be more robust to clothing variations than if it were extracted from visible images. In the mentioned works, shape information was extracted from BIOGIGA database [8], which contains synthetically generated mmW images. They exploited geometrical measures between different silhouette landmarks and features based on contour coordinates, respectively. In all cases, images were extracted in the range of 94 GHz. In a follow up work, Gonzalez-Sosa *et al.* [7], explored the potential use of shape information using a real mmW dataset, which is composed of passive images from 50 individuals. The results concluded that person recognition might be feasible, reaching performance rates of around $10\%$ EER.

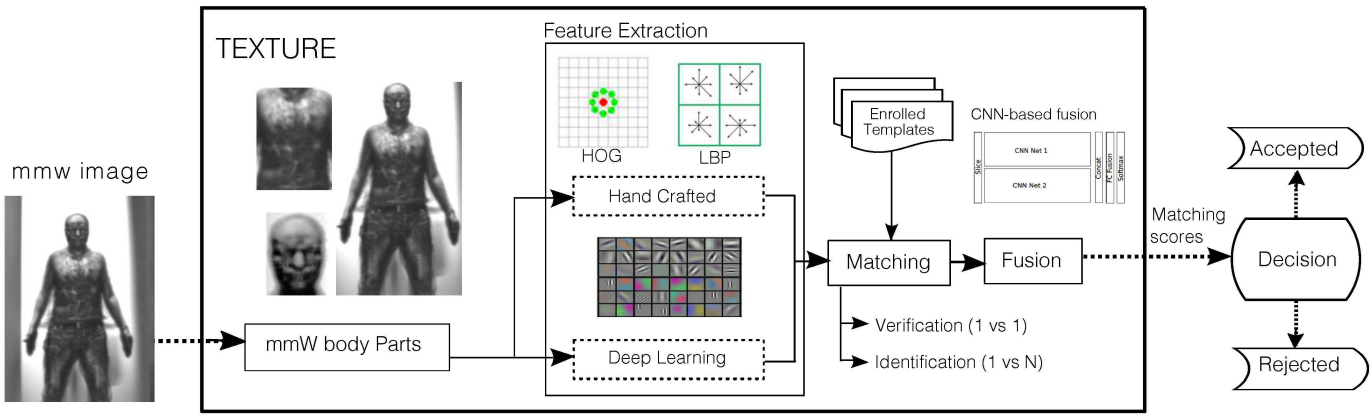

Fig. 1: **General diagram of the mmW Person Recognition system through texture**. Three mmW body parts are considered: face, torso and wholebody. Both hand-crafted and deep learning features are assessed.

In this work, we present further insight about the potential of use of mmW images for person recognition, after our first work on this line [10]. As mmW waves have the ability to pass through clothes, person recognition may be achieved not only through face information, but also through other parts of the body such as the torso or even the whole body. The main contributions of this work in comparison with our previous work [10] are the following:

- Exhaustive experimental analysis of the discriminative capabilities of body texture information from mmW images for three body parts: face, torso and the whole body. For this analysis both hand-crafted features and deep convolutional neural network (CNN) features are explored.
- Exploration of two fusion schemes, i.e., multi-algorithm and multi-modal, are proposed to further improve the recognition performance. We also propose a new multi-modal fusion approach of face and torso based on a two branch CNN architecture.
- Comparative analysis with mmW body shape information [7] and face recognition in the visible and NIR spectral bands are also performed.

Fig. 1 depicts the whole pipeline of the mmW Person Recognition system using texture information. Given a mmW image, mmW body parts are first extracted. As mentioned before, we do not only consider the face but also the torso and the wholebody. As mmW waves can pass through clothes, texture information of the torso can provide discriminative information for the task of identification. Different feature approaches are considered, based on hand-crafted or deep learning techniques. Then, different matching schemes are explored followed by some standard and novel fusion techniques.

The rest of the paper is organized as follows. Section 2 describes the selected hand-crafted and CNN features analyzed and compared throughout the paper. Section 3 describes the matching schemes followed. Section 4 presents the mmW TNO database, while Section 5 gives details about the experimental protocol followed. Experimental results are presented in Section 6. Finally, Section 7 concludes the paper with a brief summary and discussion.

## 2 FEATURE REPRESENTATION

In this work, three mmW body parts are considered: mmW face, mmW torso and mmW whole body. These mmW body parts are extracted from the original mmW images by manually defining the corresponding bounding boxes. The approximate sizes of these bounding boxes are $70 \times 90$ for mmW face; $120 \times 170$ for mmW torso and $250 \times 450$ for the body ($width \times height$ format). Then, mmW faces are histogram equalized using the INface toolbox v2.0 [11]. Millimeter torso and body are not histogram equalized.

Regarding feature representation, we consider two approaches. Features are first extracted from conventional hand-crafted approaches. We also explore various features based on deep learning that have been quite successful in object and face recognition.

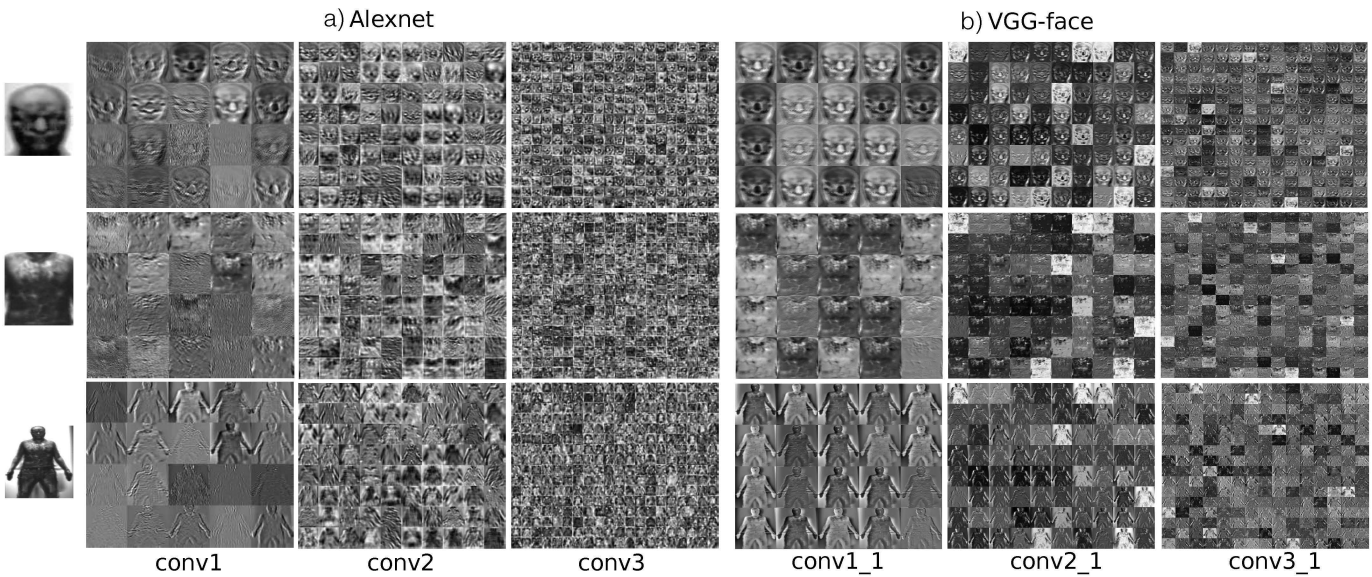

Fig. 2: Feature maps from the resulting Alexnet (a) and VGG-face (b) fine-tuned nets for the different mmW parts: face (first row), torso (second row) and whole body (last row). We show Alexnet features from conv1, conv2 and conv3 layers; while VGG-face features are shown from conv1_1, conv2_1 and conv3_1 layers. Notice that VGG-face feature maps have a higher dimension than Alexnet feature maps, but they are displayed here with similar sizes.

## 2.1 Hand-Crafted Features

Among the wide variety of hand-crafted features presented in the literature for biometric recognition, we select two of the most widely used features: $i$) Local Binary Patterns (LBP), and $ii$) Histogram of Oriented Gradients (HOG).

2.1.0.1 Local Binary Patterns features: provide discriminatory texture information that has been proved to be robust against illumination variations [12]. In this case, first, images are resized to $100 \times 150$ ($width \times height$ format). Each image is then divided into non overlapping $10 \times 10$ blocks. For each block the LBP histogram feature is computed with radius 1, 8 neighbours and uniform patterns, resulting in a 59-length vector. The final feature vector of each image is the concatenation of all the histograms from all blocks. Note that the final feature vector is in terms of histograms of LBP from each block. LBP features are extracted based on the implementation provided by [13].

2.1.0.2 Histogram of Oriented Gradients features: are able to retain both shape and texture information. Likewise, first, images are resized to the same dimension used for LBP features. HOG features are also computed block-wise and are based on histogram calculation followed by histogram normalization. Each $10 \times 10$ block is described by a histogram of gradients with 8 orientations, with each gradient quantized by its angle and weighted by its magnitude. Then, four different normalizations are computed using adjacent histograms, resulting in $8 \times 4$-length feature vector for each block. The final feature vector of a given image is the vectorization of the HOG features from all blocks. Each of the HOG features are extracted using the implementation provided in [14].

## 2.2 Convolutional Neural Network features

In recent years, features obtained using deep Convolutional Neural Networks (CNN) have yielded impressive results on various computer vision and biometrics recognition problems such as face recognition and object recognition [15]. However, not all the unsolved pattern recognition problems possess huge datasets to exploit CNN learning capabilities. Fortunately, there are ways to overcome to some extent this limitation, e.g.: transfer learning. Transfer learning is the ability of a system to recognize and apply knowledge and skills learned in previous tasks to novel tasks [16], [17]. In this definition, transfer learning aims to extract the knowledge from one or more source tasks and applies the knowledge to a target task. In the context of CNN, transfer learning can be applied using the same CNN architecture for the source and target tasks. Rather than using randomly initialized weights for training in the target, the training procedure starts with the parameters (weights and biases), estimated in the source task. This is possible because the lower layers (the closest ones to the input layer) in CNNs learn low-level features, and the layers closer to the output learn high-level features. There are two major transfer learning strategies:

- **Use the network as feature extractor:** in this case the deep network is used as a feature extractor. This can be done by feed forwarding images from the target dataset and take the activations from an intermediate hidden layer as features. The layers most commonly used for feature extraction are the first fully connected layers after the conv-relu layers. Once we extract this feature vector for all images from the target dataset, a classifier can be used (e.g. SVM or Softmax classifier) for classification.
- **Fine-tune the network:** The second strategy is focused on adjusting the weights (fine-tune) from the pre-trained

network that optimizes the new target classification problem, by continuing the stochastic gradient descent. In this regard, there are also two alternatives $i)$ fine-tune the whole network, or $ii)$ freeze the earlier layers by setting their learning factor to 0 (to prevent overfitting), and only fine-tune some higher-level portion of the network, normally the last three layers. This is motivated by the observation that the earlier features of a CNN architecture contain more generic features that could be useful to many tasks (e.g. source and target tasks) but later layers of the network becomes progressively more specific around the details of the classes contained in the original dataset.

Here we use two pretrained CNN models, namely AlexNet [18] and VGG-face [19] to extract features for mmW person recognition using a subset of the mmW TNO dataset. We analyze both transfer learning techniques: feature extractor and fine-tuning. In this case the source task is object recognition for Alexnet and face recognition for VGG-face and the target task is person recognition through mmW texture information using the mmW TNO as the target dataset. For more details regarding the training parameters, see [10].

Table 1 describes the configuration of both CNNs. As can be seen, VGG-face has a larger number of layers, hence a larger number of learnable parameters. Notice also from both networks that as it goes deep in the network, the number of filters ($depth$ dimension) in a convolutional layer increases (see e.g. for Alexnet 96, 256 and 384 filters for conv1, conv2 and conv3, respectively).

While it is true that the input size of both networks is very similar, the output size of the first three convolutional layers are quite different between Alexnet and VGG-face. As the latter contains more conv-relu layers, it may reduce the dimensionality of convolutional outputs more gradually than in Alexnet. Fig. 2 a) and b) depicts the outputs or feature maps of conv1, conv2 and conv3 layers from Alexnet and conv1_1, conv2_1 and conv3_1 from VGG-face respectively, for the mmW face, mmW torso and mmW whole body. As can be seen from this figure, the layers closest to the input layer contain a lower number of filters with low-level features such as edges. Conversely, deeper convolutional layers hold a larger number of filters but with a lower size, addressing high-level features such as complex shapes or fine details.

## 3 MATCHING

We explore two different matching schemes: cosine similarity and Softmax.

TABLE 1: Alexnet and VGG-face configurations. Size follows $width \times height \times depth$ format.

| Description | Alexnet [18] | VGG-face [19] |
|---|---|---|
| # of layers | 21 | 39 |
| # of conv-relu layers | 5 | 16 |
| # of parameters | 60 M | 135 M |
| Input Size | $227 \times 227 \times 3$ | $224 \times 224 \times 3$ |
| Output Size of conv1 | $55 \times 55 \times 96$ | $224 \times 224 \times 64$ |
| Output Size of conv2 | $27 \times 27 \times 256$ | $112 \times 112 \times 128$ |
| Output Size of conv3 | $13 \times 13 \times 384$ | $56 \times 56 \times 256$ |

3.0.0.1 Cosine Similarity: is the distance-based matcher selected for computing match scores between either hand-crafted or deep features, defined as:

$$\text{CS} = \frac{\mathbf{a}^T \cdot \mathbf{b}}{||\mathbf{a}|| \cdot ||\mathbf{b}||}, \quad (1)$$

where $\mathbf{a}$ and $\mathbf{b}$ are the feature vectors to be compared (column vectors and $T$ denotes transpose).

3.0.0.2 Softmax classifier: generalizes the Logistic regression classifier to multiple classes. The softmax function is the supervised classifier used in the final layers of CNN architectures only for identification mode. Given $N$ different subjects, the probability that the predicted class $y$ will be assigned to subject $j$ given a particular feature vector $\mathbf{x}$, is computed as:

$$P(y=j|\mathbf{x}) = \frac{e^{\mathbf{x}^T \mathbf{w}_j}}{\sum_{k=1}^{N} e^{\mathbf{x}^T \mathbf{w}_k}}. \quad (2)$$

In this particular case $\mathbf{x}$ are referred to the outputs of the last fully connected layers of the CNN architecture and $\mathbf{w}_k$ with $k = 1, \ldots, N$ are the softmax classifier parameters.

### 3.1 Fusion Schemes

In Biometrics, fusion schemes emerge as an avenue to, among other reasons, improve the performance and robustness of individual biometric systems [20]. The idea relies on combining information coming from different sources. In what follows, we refer to multimodal fusion approaches when sources of information come from different biometric traits and to multi-algorithm fusion approaches when those sources come from the same biometric trait but using different feature extraction algorithms. For instance a multimodal fusion approach fuses information from face and fingerprint while a multi-algorithm fusion approach combines pixel values and LBP values extracted from the face biometric trait.

There are different strategies to combine information from different sources. Concretely, information fusion can be carried out at each of the different levels in which a biometric system is usually divided (sensor, feature, score and decision):

3.1.0.1 Feature level fusion: refers to the combination of different feature vectors. These feature vectors may be obtained from different sensors or by applying different feature extraction algorithms to the same biometric data. In the context of this research, the fused feature vector results from the concatenation of individual feature vectors. In this work feature level fusion is only analyzed with multimodal approaches. For instance, the feature level fusion of LBP extracted from mmW faces and mmW torso results in a 17700-length feature vector (each mmW body part is described by 10 blocks $\times$ 15 blocks $\times$ 59-LBP histogram values $=$ 8850$-$length vector).

3.1.0.2 Score level fusion: refers to the combination of the different scores provided by the different systems involved. In this particular case, scores are fused using the *sum* fusion. This information fusion strategy is used for both multimodal and multi-algorithm fusion approaches. Notice that scores have not been normalized, as matchers from individual systems are already in the same range $[0, 1]$.

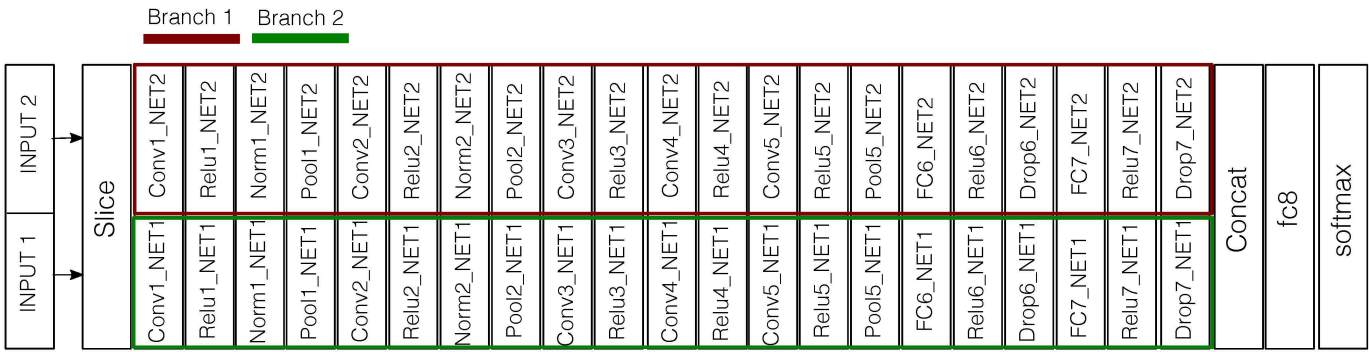

Fig. 3: **CNN level fusion**. The multimodal architecture for mmW Person Recognition with the Alexnet pre-trained model.

3.1.0.3 CNN level fusion: is inspired by previous works in the area of object recognition [21]. Here we explore this architecture to perform person recognition using information from different body parts within a single CNN architecture. Concretely, this multimodal network consists of two CNN branches, which are then combined in a late fusion approach. The idea behind this architecture aims to jointly learn discriminative features from different body parts, for instance face and torso. This CNN level fusion has been carried out for both pre-trained models considered: Alexnet and VGG-face. Fig. 3 depicts the overall architecture of the proposed CNN level fusion, using the Alexnet architecture. As can be seen, branch 1 and branch 2 are merged by concatenating features from the $fc7$ layer, resulting in a concat vector of size $2 \times 4096 = 8192$. After that, an additional fully connected layer is introduced followed by the *soft max* classification layer. To assure that mmW body parts come from the same identity, a *slice* layer is introduced at the beginning of the network. This fusion scheme is only feasible for multimodal fusion approaches. Notice also that the training of this multimodal CNN is performed stage-wise, that is, first each branch is trained individually and then the overall architecture is trained using the individual trained models from branch 1 and branch 2. Due to the lack of a sufficient number of samples, both branches are fine-tuned using a particular pre-trained model (Alexnet or VGG-face). After that, the overall CNN architecture is trained to learn the parameters associated to the new introduced layer $fc8$, using a batch size of 50, learning rate of 0.0001 and 100 epochs.

## 4 THE MMW TNO DATABASE

The mmW TNO database (created by the Dutch Research Institute TNO in The Hague) is one of the very few resources for research that contains real images of subjects extracted in the range of mmW specifically designed for person recognition purposes [6]. Images were recorded using a passive stereo radiometer scanner in an outdoor scenario.

The database is comprised of images belonging to 50 different male subjects in 4 different scenarios. These 4 different scenarios derive from the combination of 2 different head poses and 2 different facial occlusions. In the first head pose configuration, the subject is first asked to stand in front of the scanner with head and arms position fixed (*frontal head pose*). In the second pose configuration (*lateral head pose*), the subject is asked to turned his head leftward while the torso is asked to remain fixed (it may suffer some small changes due to the head movement). A second round of images with the first and second head pose configurations were extracted but now a large part of the facial region was occluded using an artificial beard or balaclava (disguised). For more details regarding the image acquisition and the hardware employed, we refer the reader to the original paper [6].

As mentioned before, each scanning is a set of two grayscale images. By dividing this set into single images of $348 \times 499$, the TNO database is comprised of 50 subjects $\times$ 2 head pose configurations $\times$ 2 facial clutter configurations $\times$ 2 images per set, making a total of 400 images in the whole mmW TNO database. A sample $348 \times 499$ image from this dataset is shown in Fig. 4.

## 5 EXPERIMENTAL PROTOCOL

The experimental protocol followed in the previous work using the mmW TNO database was very optimistic [6]. It assumed 4 images as input (test images) and 4 images as training. Individual distances were computed comparing pairs of images under the same head pose and point of view conditions, which is not a realistic situation. Then, the final distance was the minimum over the 4 former individual distances.

Subjects were scanned following a cooperative protocol with the arms upwards and legs separated between each other. As the idea is to study the possibility of using the same mmW scanned images for both concealed weapon detection and person recognition applications, we elaborate our study under the assumption of this cooperative protocol.

The experimental protocol proposed in this work aims to simulate the real situation in which a traveler would enter in the mmW scanner deployed in the security area of an airport. Enrollment is carried out in the first use of the mmW scanner. At the same time the subject is being scanned to target concealed weapons or dangerous objects, he is also compared with the previously enrolled template associated to the identity claimed in his passport. To simulate this scenario, we report results in verification mode. It would also be possible to compare the mmW image of the person with a watchlist of suspects. This case has been also considered and identification results are also reported.

Additionally, in order to gain insight of the benefits of using mmW images, we explore different experimental protocols:

TABLE 2: **Individual Results for Verification**. Results are reported in terms of EER (%) for all feature approaches and the three mmW body parts. Abbreviations used: Histogram of Oriented Gradients (HOG); Local Binary Patterns (LBP), Feature Extractor (Feat. Ext).

| mmW Body Part | HOG | LBP | Alexnet Feat. Ext. | Alexnet Fine Tune | VGG Feat. Ext | VGG Fine Tune |
|---|---|---|---|---|---|---|
| Face | 35.50 | 33.00 | 31.5 | 25.50 | 28.5 | 27.00 |
| Torso | 4.50 | 6.00 | 11.46 | 6.00 | 14.34 | 8.72 |
| Wholebody | 34.00 | 35.50 | 9.00 | 14.00 | 17.30 | 9.50 |

TABLE 3: **Individual Results for Identification**. Results are reported in terms of rank one identification rate (R1) in % for all feature approaches and the three mmW body parts. Abbreviations used: Histogram of Oriented Gradients (HOG); Local Binary Patterns (LBP), Feature Extractor (Feat. Ext).

| mmW Body Part | HOG | LBP | Alexnet Feat. Ext. | Alexnet Fine Tune | VGG Feat. Ext. | VGG Fine Tune |
|---|---|---|---|---|---|---|
| Face | 29 | 31 | 22 | 40 | 56 | 49 |
| Torso | **98** | 97 | 87 | 91 | 78 | 88 |
| Wholebody | 55 | 37 | 93 | 88 | 80 | 80 |

5.0.0.1 Frontal protocol.: This protocol is set up as the baseline protocol. Among the 4 images with *frontal head pose*, we randomly select 2 images as gallery images and 2 images as probe images.

5.0.0.2 Cross-pose protocol.: With this protocol, we aim to study the robustness of the different shape-based features proposed against pose variations. The mmW TNO database contains images with *frontal head pose* and with *lateral head pose* (see Fig. 4). To this aim, we randomly select 2 images from the *frontal head pose* subset as gallery images, and 2 images from the *lateral head pose* subset as probe images.

We report experiments in both verification and identification modes.

### 5.1 Verification Mode

Verification implies one-to-one comparisons to find out whether the two given images belong to the same subject or not. In this case we face 100 training images (2 samples of 50 subjects) against another set of 100 testing images. Hand-crafted features are extracted as mentioned in Section 2.1. In order to extract the CNN features, we feed forward both training and test images in the corresponding fine-tuned networks and extract the features from the next-to-last fully connected layer, resulting in a 4096-feature vector (fc7).

Finally, for both hand-crafted and CNN features, training features are matched against test features using cosine distance to obtain genuine and impostor scores, yielding 100 genuine scores and 4900 impostor scores.

### 5.2 Identification Mode

In the identification mode, a model is learned for each subject in the dataset. The template of the hand-crafted features is computed by averaging the features from the 2 images belonging to the same class (subject).

In order to get identification results, all test images are compared to the 50 subject templates. For the hand-crafted features, we compute cosine distances between a particular subject model and the test features. For CNN, we just feed forward the corresponding fine-tuned network (Alexnet or VGG-face) with the test images to get the class scores from the classification layer.

## 6 RESULTS

Here we present the performance results for both verification and identification modes. Identification results are reported in terms of rank one identification rate (R1), while verification results are reported in terms of EER. First we analyze the proposed texture approaches: two hand-crafted and two CNN-based. All the approaches are tested with the three considered mmW body parts: mmW face, mmW torso and mmW whole body. Two fusion schemes, i.e., multi-algorithm and multi-modal, are proposed to improve the recognition performance. Later, a comparison of shape and texture information for mmW body images is reported. Finally, a comparison of face recognition for the visible, near infrared (NIR) and mmW spectral bands is also given.

### 6.1 Body Parts

Table 2 reports the individual results achieved by the three mmW body parts and the four different feature approaches. Results using CNN are reported using both transfer learning techniques explained in Section 2.2: feature extractor and fine tuning. In this regard, we observe in general a superior performance of deep learning techniques while using the fine tuning strategy, i.e. adapting the very last layers of the CNN architecture to the new task. We think this can be due to the several dissimilarities between images from source and target datasets: $i$) images from source datasets are in RGB while images from the target dataset are in grayscale; $ii$) images from source datasets are based on natural images for Alexnet, or faces for VGG-face, while the target dataset includes faces, torsos, and wholebody; and $iii$) images from source and target datasets belong to different spectrum: visible (source) and the mmW (target), having also different resolution.

It can be observed from Table 2 that person recognition using mmW faces achieves around 30% EER for all approaches, which are far from being comparable to the state-of-the-art results in the visible spectral band (with results around 99% of accuracy on LFW dataset [22]). CNN features are able to reach results slightly better than hand-crafted features, obtaining a 27% of relative improvement between the average EER of hand-crafted and CNN features. One of the reasons that could explain this poor performance is that the resolution associated to mmW images (lower than with visible images) is insufficient to obtain discriminative information from facial mmW images. It is also worth noting the importance of applying histogram equalization to mmW faces. We empirically proved a reduction of EER by an average of 7% for the different approaches when using histogram equalized images.

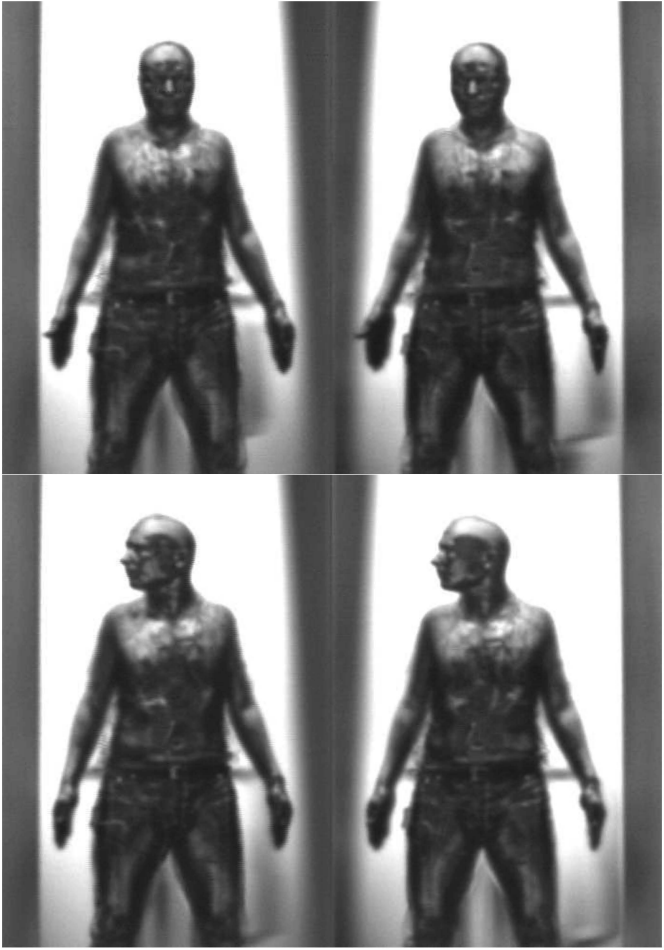

Fig. 4: Two screening images from the mmW TNO Database acquired with a stereo radiometric scanner. A screening image is comprised of two single images with slight variations in pose. Two different screening images are shown with different viewpoints: frontal (left) and lateral (right). Figures are extracted from [6].

We also observe that millimeter torsos provide rich discriminative information for person recognition. Superior performance of mmW torso can be seen from Table 2, in which all considered features achieve outstanding results, reaching the best result of 4.5% EER with HOG features. This may be due to the fact that torso images are the most stable region across the database as this region was extracted based on key points of the body silhouette. We believe texture is providing the most discriminative information since when using exclusively shape torso information, the EER degrades up to 17.97% (reported in our previous work [23]). Face images are much smaller in size and present a much larger degree of variability in terms of pose or expression changes. In addition, whole body images present many variabilities regarding the positioning of the arms and legs in each image, which can lead to a lower recognition performance. Also, the fact that lower body clothes material are thicker (e.g. jeans), mmW waves penetration is lower and therefore the useful information from that part is reduced.

Hand-crafted features with mmW whole body are performing similarly as with mmW faces. In this case, a previous alignment of the body may have helped to improve performance of these spatial-dependent features. CNN features are performing reasonably well, being more discriminative than face but less than torsos. It is interesting to note that the VGG-face features perform slightly better than the Alexnet features. Notice that the VGG pre-trained model has been trained with faces whereas Alexnet has been trained with natural images. This fact might produce a more effective knowledge transfer from VGG to the new task (mmW face recognition).

Table 3 presents the individual R1 for all mmW body parts and feature approaches. R1 indicates the percentage of probe samples where the system has assigned the right identity in the first place among the 50 possible candidates.

Similar conclusions can be drawn from the identification experiments compared to the verification experiments: the best performance of CNN features over hand-crafted features for mmW faces and mmW whole body, the superiority of torso over face and whole body regardless of the feature approach, being hand-crafted features also superior here. The best torso performance of hand-crafted features for either verification and identification may be due to the aforementioned lack of samples per class for properly fine-tuning pre-trained models, but also due to the distance between source and target dataset. Bear in mind that Alexnet has been trained using images of objects and VGG-face has been trained using images of visible faces.

It is also worth noting that not always approaches that minimize EERs in verification do also maximize R1 in identification. Such is the case of VGG-face and Alexnet for mmW faces, in which VGG-face achieves a better rank-1 (49%) than Alexnet (40%), although Alexnet achieve less EER in verification mode. This is happening also the other way around between VGG-face and Alexnet for mmW whole body. Now the approach with a larger EER achieves better R1 results. Previuos works have stated the same conclusion [24].

Regarding the computational cost, it is worth noting that the experiments have been run on Matlab R2017a on Windows 10 using a Intel Core i7-5820K processor with CPU at 3.30GHz, 32 GB RAM and a NVIDIA GeForce GTX 1080 GPU card. The computational cost of one comparison of a pair of images including feature extraction is: 0.27 sec. for LBP, 0.12 sec. for HOG, 0.029 sec. for VGG-face and 0.0058 sec. for Alexnet. These results show that deep learning implementations are very fast at test time.

TABLE 4: **Cross-pose protocol.** Impact of pose variations over mmW torsos for verification and identification modes. Results are reported in terms of EER (%) and R1 (%), respectively.

| System | Frontal Protocol | | Cross-pose Protocol | |
|---|---|---|---|---|
| | EER | R1 | EER | R1 |
| HOG | 4.50 | 98 | 9.97 | 95 |
| LBP | 6.00 | 97 | 12.50 | 92 |
| Alexnet | 6.00 | 91 | 10.00 | 81 |
| VGG | 8.72 | 88 | 15.00 | 75 |

## 6.2 Body Position

In this section, we aim to gain insight about the robustness of the different proposed methods when the body position varies.

Table 4 reports verification and identification results achieved when following the cross-pose protocol with mmW torso, which is the mmW body part with best performance. As can be seen, pose variation degrades performance slightly and similarly for all texture-based feature approaches considered. Hand-crafted features keep their superior performance above deep learning features.

## 6.3 Fusion Schemes

### 6.3.1 Multi-algorithm Fusion

Table 5 and Table 6 show the results achieved after performing multi-algorithm fusion for identification and verification modes, respectively. As stated before, the multi-algorithm fusion is performed at score level. In both cases, three different fusions are considered: *Fusion ALL* in which all texture approached are considered; *Fusion HC* in which only Hand-Crafted features are taken into account (HOG and LBP) and *Fusion DL* where only Deep Learning features are combined (Alexnet and VGG).

As can be seen from Table 5, person recognition through mmW face benefits more when considering the whole set of texture features, reaching an improved $R1$ of 54%. Considering mmW torso, none of the fusion schemes is able to outperform the best individual $R1$ achieved with HOG features. In the case of mmW whole body, the improvement reached while considering exclusively features from Deep Learning approaches suggest that features can complement each other if their corresponding individual performance are alike. When there is a considerable gap of performance, it is better to discard features with worse performance.

Similar conclusions can be drawn from multi-algorithm fusion in verification mode. As observed in Table 6, the best fusion results combining different texture approaches for mmW face and mmW whole body are achieved when only considering deep learning features. This is because there is a notable difference in individual performance between deep and hand-crafted approaches. Contrary to mmW face and mmW whole body, person recognition through mmW torso achieved the best performance when all texture approaches are considered. This is because all texture approaches perform fairly well and hence it is more likely that features can complement each other.

### 6.3.2 Multimodal Fusion

Table 7 and Table 8 give account of the multimodal fusion results achieved for identification and verification, respectively. In both cases, multimodal fusion is applied at feature, score and CNN level. We explore the following combination of biometric traits: face-torso, torso-whole body and face-whole body, using a particular feature approach: LBP, HOG, Alexnet and VGG. Note that for LBP and HOG, only feature and score level are possible. We will proceed to discuss first the multimodal fusion results attained for identification and then for verification.

As can be seen from Table 7, when using LBP features, fusion carried out at score level attains considerable better results than the counterpart fusion performed at feature level. Despite that, none of the multimodal fusion configurations using LBP is able to outperform the best individual system, which is attained by the torso with R1 of 97%. Conversely, multimodal fusion results attained with HOG features prove that feature level fusion scheme is slightly more appropriate than score level fusion. In this case, a feature level fusion between torso and whole body modalities manages to improve mildly the best individual results, increasing $R1$ of 98% attained with the torso up to 99% with the multimodal approach.

In what concerns multimodal approaches using deep learning features, it is observed that in most cases, multimodal fusion approaches outperform individual performances of the modalities considered.

It is also worth nothing that performances attained at feature and CNN level are very similar, as Table 7 shows. The main reason for this is that in fact the CNN level fusion is a type of feature level fusion as it concatenates features at the $fc7$ layer. Actually, the main difference between these two fusion approaches for the deep learning approaches relies on the classifier employed, being cosine distance for feature level and soft max for CNN level. Even if results are very similar, in general CNN level reaches slightly better results than feature level. Concretely, face and torso multimodal fusion reaches their best performance when fusion is carried out at CNN level for both Alexnet and VGG-face. Torso and whole body configuration obtains optimal results with CNN level for Alexnet and with feature-level for VGG. On the other hand, face and whole body reach their best fusion performance when fusing information at score level. It should be noted that the best performance is achieved when: $i)$ fusing HOG features from torso and whole body at feature level, or $ii)$ fusing Alexnet features from torso and whole body at CNN level. Table 8 gives account of the fusion results attained for verification which are in line with the ones obtained for identification. Fig. 5 a) and b) depicts the best fusion schemes achieved for both verification and identification modes respectively following both multi-algorithmic and multimodal fusion schemes.

## 6.4 Comparative Analysis of mmW Body Shape and Texture Information

Now that we have analyzed the recognition performance of mmW texture information for the three body parts considered, it is worth comparing these results with those achieved using only the body shape information for the same images. In a recent work [23], we carried out a deep analysis on using body shape information from mmW images for person recognition. Different feature approaches (in particular Contour Coordinates, Shape Contexts, Row and Column Profiles and Fourier Descriptors) were considered with three matching approaches (Dynamic Time Warping, Modified Hausdorff Distance and Support Vector Machines). The experimental protocol followed was exactly the same as the one followed in this work, allowing for a comparative analysis between body shape and body texture information for person recognition.

TABLE 5: **Multi-algorithm fusion in identification mode**. Results are reported in terms of R1 (%) for all feature approaches and the three mmW body parts. Abbreviations used: Hand-crafted (HC); Deep Learning (DL).

| mmW Body Part | HOG | LBP | Alexnet | VGG | Fusion ALL | Fusion HC | Fusion DL |
|---|---|---|---|---|---|---|---|
| Face | 29 | 31 | 40 | 49 | **54** | 36 | 52 |
| Torso | 98 | 97 | 91 | 88 | 97 | **98** | 93 |
| Wholebody | 55 | 37 | 88 | 80 | 94 | 45 | **95** |

TABLE 6: **Multi-algorithm fusion in verification**. Results are reported in terms of EER (%) for all feature approaches and the three mmW body parts. Abbreviations used: Hand-crafted (HC); Deep Learning (DL).

| mmW Body Part | HOG | LBP | Alexnet | VGG | Fusion ALL | Fusion HC | Fusion DL |
|---|---|---|---|---|---|---|---|
| Face | 35.50 | 33.00 | 25.50 | 27.00 | 25.00 | 33.50 | **22.00** |
| Torso | 4.50 | 6.00 | 6.00 | 8.72 | **2.00** | 4.00 | 4.50 |
| Wholebody | 34.00 | 35.50 | 14.00 | 9.50 | 20.42 | 34.50 | **9.20** |

TABLE 7: **Multimodal fusion schemes: identification**. Fusion performance in terms of rank-1(%) for all combination of two mmW body parts. Fusion is performed at three levels: feature, score and convolutional neural network. Abbreviations used: Face (F); Torso (T); Wholebody (W); Not Available (N/A); Local Binary Patterns (LBP); Histogram of Oriented Gradients (HOG). The best performance for each feature approach is bolded.

| System | Individual | | | Feature Level | | | Score Level | | | CNN level | | |
|---|---|---|---|---|---|---|---|---|---|---|---|---|
| | F | T | W | F T | T W | F W | F T | T W | F W | F T | T W | F W |
| LBP | 31 | **97** | 37 | 86 | 69 | 31 | 96 | 91 | 42 | | N/A | |
| HOG | 29 | 98 | 55 | 98 | **99** | 56 | 98 | 90 | 53 | | N/A | |
| Alexnet | 40 | 91 | 88 | 92 | 97 | 71 | 89 | 97 | 94 | 95 | **99** | 71 |
| VGG | 49 | 88 | 80 | 76 | **98** | 71 | 92 | 94 | 90 | 91 | 96 | 78 |

TABLE 8: **Multimodal fusion schemes: verification**. Fusion performance in terms of EER (%) for all combination of two mmW body parts. Fusion is performed at three levels: feature, score and convolutional neural network. Abbreviations used: Face (F); Torso (T); Wholebody (W); Not Available (N/A); Local Binary Patterns (LBP); Histogram of Oriented Gradients (HOG). The best performance for each feature approach is bolded.

| System | Individual | | | Feature Level | | | Score Level | | | CNN level | | |
|---|---|---|---|---|---|---|---|---|---|---|---|---|
| | F | T | W | F T | T W | F W | F T | T W | F W | F T | T W | F W |
| LBP | 33.0 | **6.0** | 35.5 | 16.5 | 24.5 | 33.0 | 10.5 | 16.0 | 30.5 | | N/A | |
| HOG | 35.5 | **4.50** | 34.0 | 8.4 | 9.9 | 28.3 | 9.5 | 11.6 | 29.5 | | N/A | |
| Alexnet | 25.5 | 6.00 | 14.0 | 8.5 | 5.9 | 15.3 | 6.0 | 5.5 | 13.4 | 5.5 | **2.5** | 17.0 |
| VGG | 27.0 | 8.7 | 9.5 | 13.9 | 4.5 | 14.5 | 10.6 | 4.3 | 10.5 | 7.0 | **3.5** | 13.4 |

Verification performance results in the range of 10% EER were achieved in general for the different combination of shape features and matching approaches. The best result of 8% EER was achieved when Row and Column Profile features were combined with SVM. Regarding mmW texture information, the best verification performance is achieved for the torso part with a fusion of hand crafted and deep features obtaining an EER of 2%.

TABLE 9: **Comparison of mmW body shape and body texture approaches.** Verification performance in terms of EER (%) and identification performance in terms of R1 (%).

| Mode | Body Shape [23] | Body Texture |
|---|---|---|
| Verification (EER) | 8.00 | 2.00 |
| Identification (R1) | 90.50 | 99.00 |

Table 9 shows the comparative results of mmW body shape and body texture for both verification and identification modes. For the identification mode, the best Rank-1 accuracy of 90.5% was achieved for the shape approach comprised of Contour Coordinate features with Modified Hausdorff Distance. The best identification performance based on the mmW texture information is achieved using the multimodal fusion of torso and the whole body with feature level fusion of HOG features with Alexnet CNN features obtaining the Rank-1 identification rate of 99%. In both cases the performance achieved by mmW body texture approaches is significantly better. A reason for this is that mmW body texture approaches used in this paper also extract and consider information of the body shape, therefore it is reasonable to achieve a better recognition performance. In our future work, an exploration of the fusion of mmW body shape and texture approaches will be carried out.

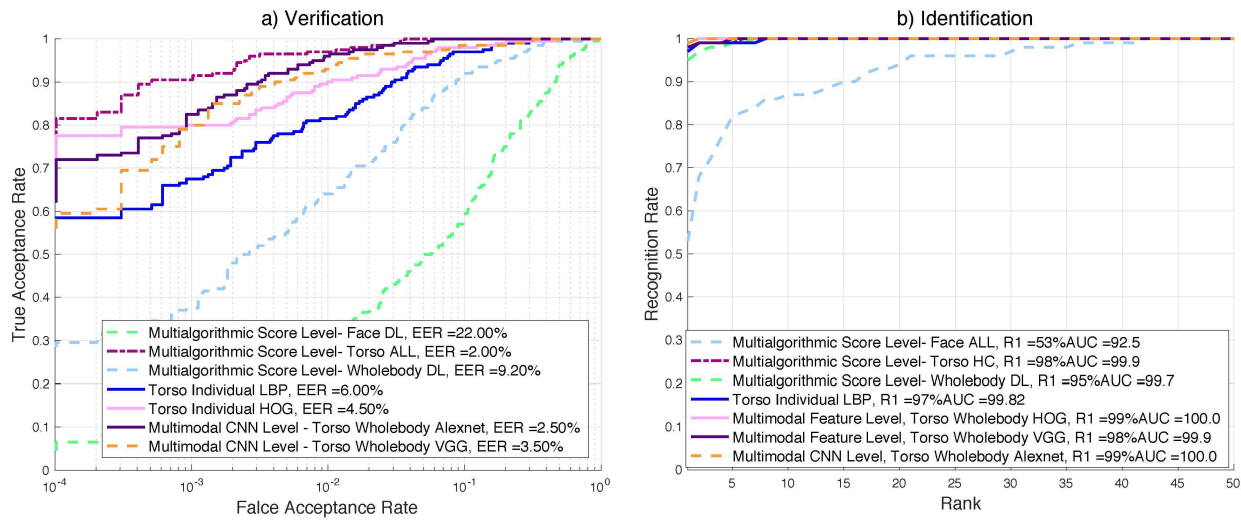

Fig. 5: **Best Fusion Results**. Best Fusion Results achieved in verification mode (a) and identification mode (b). In each case we plot: $i)$ the best multialgorithmic fusion scheme achieved by each mmW body part (face, torso, and wholebody), $ii)$ the best multimodal scheme for each feature approach (LBP, HOG, VGG and Alexnet). Abbreviations used: Deep Learning (DL); Hand Crafted (HC); ALL (All feature approaches).

### 6.5 Face Recognition in Different Spectral Bands

Motivated by the low recognition performance achieved by the mmW texture information from face images, we report here a comparative analysis of face recognition over other two spectral bands: visible light (VIS) and near infrared (NIR), applying the same methods proposed in this work, i.e., hand crafted features, deep features and their fusion.

Fig. 6 shows a diagram of the electromagnetic spectrum. In this paper, we have focused on the mmW band due to their good properties such as their ability to pass through clothing allowing the detection of concealed weapons and person recognition, as the results of this work show. In the application scenario we consider in this work, i.e., the person enters a mmW scanner at an airport, there are not constrains for not using other ranges of the spectrum as well. A mmW scanner could also incorporate visible light or near infrared cameras to acquire high quality face images.

In the last years, there have been many research works on face recognition in the infrared (IR) band [26], [27], due to some good properties compared to the visible band. In particular, research has been conducted in the following subbands: 1) intensified near infrared (NIR), which is robust to illumination and able to provide visibility in low-light conditions [25], [28]; 2) shortwave IR (SWIR), which is more tolerant to low levels of obscurants like fog and smoke compared to NIR [29], [30]; 3) middle wave IR (MWIR), which is invariant to illumination and acquires temperature variations across the face at a distance [29], [31]; and 4) long wave IR (LWIR), with similar properties to MWIR, but revealing different image characteristics of the facial skin [32]. Some of these works are focused on multispectral face matching [25], [29], [30], i.e., matching face images from different spectral bands (e.g., visible to NIR).

In order to compare the methods proposed in this work for mmW face verification with other spectral bands, we have used the CASIA NIR-VIS 2.0 face database [25]. This database is comprised of a total of 725 subjects with 1-22 visible (VIS) and 5-50 NIR face images per subject. Fig. 7 shows some example of visible and NIR images considered here. As the experimental protocol for this database was focused on multispectral face matching, we decided to divide the database into two main sets, one for system development comprised of 400 subjects with 6 available images for both VIS and NIR bands, and another one for system evaluation consisting of the remaining 325 subjects with a variable number of images per subject. Deep learning-based methods were fine-tuned using the development dataset. Then, the evaluation performance results for both VIS and NIR images were computed over the evaluation dataset using the cosine distance over pairs of face images for both deep learning and hand-crafted approaches. The fusion of the feature approaches was also computed at the score level. In the evaluation, a total of 2180 genuine and 657622 impostor scores were computed. The same configuration and parameters of the methods proposed in this paper were used in this experiment for a fair comparison.

Table 10 shows the EER verification results achieved for the different feature approaches and their fusion for the visible, NIR and mmW face images. Results for mmW face verification are taken from Table 6. It is interesting to see such a different performance for the three spectral bands, obtaining worse performance as the wavelength increases. Face recognition in the visible spectral band achieves the best results with very low EER values for both hand-crafted and deep features. The best result in this case is achieved as the fusion of all feature approaches obtaining 1.61% EER. These results are in line with state-of-the-art results achieved visible face recognition using deep learning approaches [22]. For the case of NIR face images there is a significantly better performance corresponding to deep features compared to hand crafted features. In particular, the best result is achieved for the fusion of the deep features obtaining an EER of 10.40%. Finally, the best result for mmW face images was also achieved for the fusion of the deep learning approaches obtaining an EER of 22.00%. The main reason for such a significant difference of performance may be due to the resolution of the images, being the images in the visible band very sharp and controlled, as the example shown in Fig. 7. From Fig. 7, one can see that NIR images have a lower resolution with blurry edges. Finally, mmW images are very noisy with not many details and have a very low resolution, as can be seen from Fig. 4. Another reason for obtaining worse performance when increasing the wavelength in deep learning-based approaches is that Alexnet and VGG models are trained using the VIS images, which are very different from the NIR and mmW images.

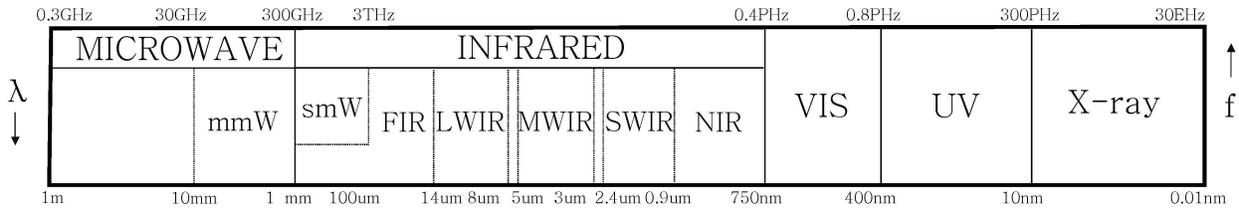

Fig. 6: Ranges of the electromagnetic spectrum.

TABLE 10: **Comparison of face recognition performance for VIS, NIR and mmW spectral bands**. Results are reported in terms of EER (%) for all feature approaches and the three spectral bands. Abbreviations used: Hand-crafted (HC); Deep Learning (DL).

| Spectral Band | HOG | LBP | Alexnet | VGG | Fusion ALL | Fusion HC | Fusion DL |
|---|---|---|---|---|---|---|---|
| Visible | 3.58 | 2.38 | 3.39 | 3.67 | **1.61** | 2.57 | 2.27 |
| NIR | 21.19 | 24.70 | 13.02 | 11.05 | 14.45 | 23.53 | **10.40** |
| mmW | 35.50 | 33.00 | 25.50 | 27.00 | 25.00 | 33.50 | **22.00** |

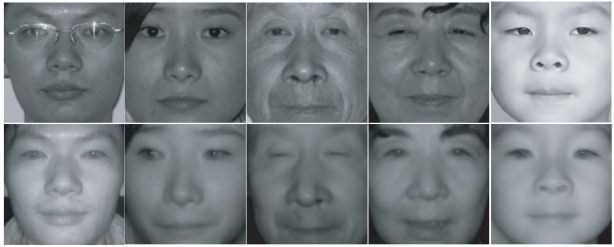

Fig. 7: Cropped VIS (top row) and NIR (bottom row) face images from the NIR-VIS 2.0 database. Each colum represents one person. Extracted from [25].

Having in mind these results, facial mmW images could be very useful for detecting artificial occlusions and artefacts that could be used for presentation attacks, but they would not be very suitable for person recognition, where the usage of a visible light facial camera would be more appropriate. Moreover, improved person recognition performance could be achieved after fusing mmW torso and/or mmW whole body images with visible face images.

## 7 CONCLUSION

This work has presented one of the very first works addressing person recognition using texture extracted from mmW images. Different mmW body parts have been considered: face, torso and whole body. We have carried out experiments with several hand-crafted features and some state-of-the-art deep learning features. Some of the findings from the experiments are: $i$) mmW torso is the most discriminative body part, followed by whole body and face with average EER of 6.30%, 23.25%, and 29.75%, respectively, $ii$) mmW face recognition has very low performance, $iii$) CNN features overcome hand-crafted features with faces and whole body parts and, $iv$) hand-crafted features achieve outstanding results for torso-based person recognition.

Finding $ii$) may be explained by the low resolution of mmW images. Comparative results of face recognition for visible, NIR and mmW images reported show how the methods proposed in this paper achieve very high recognition performance for high resolution face images in the visible region. We believe that this problem could be overcome by addressing it as a low resolution face recognition problem, which has been studied deeply elsewhere [33]. The most common techniques are based on super resolution [34] or face hallucination [35], [36].

The second experimental part of this work has addressed fusion schemes for multimodal and multi-algorithm approaches. In what concerns multi-algorithm fusion, in general, it is convenient to fuse information from all feature approaches only if individual performances are similar, otherwise it is better to discard the feature approaches with worse performance. Regarding multimodal approaches, the best fusion approach for LBP and HOG features is at score and feature level, respectively. However, none of the fusion approaches is able to outperform the best individual performance achieved with mmW torso. With learned features, CNN level fusion outperforms feature and score level approaches. In this case, CNN level fusion is able to attain up to 99% of R1 or 2.5% of EER, when using torso and whole body mmW body parts.

The best verification results for multi-algorithm fusion is obtained combining all different texture features extracted from the torso with an EER of 2%. Regarding multimodal fusion, the best result is obtained when fusing mmW torso and mmW whole body at CNN level with Alexnet, reaching a 2.5% of EER. Also, for future work, we will consider fusing texture with shape (see previous work [7] and texture information jointly [10]). Also, it would be worth exploring the possibilities of combining mmW and images from other ranges of the spectrum [26]. In particular, the results reported would suggest a fusion of mmW torso and/or the whole body with visible face images would achieve improved recognition results.

We believe one of the main limitations of this work is the relatively small size of the database, containing 50 subjects. In a real setting considering a larger number of subjects it is likely that the identification rates would vary as rank 1 and rank 5 identification rates depend on the number of subjects considered. Verification rates are more stable as they are based on one-to-one comparisons, which do not depend so directly on the number of subjects. On the other hand, it is important to notice that a larger database would allow us to better exploit the learning capabilities of the CNNs, therefore, making possible to achieve even better recognition rates. To overcome this limitation, we would need larger datasets of mmW images which is a real challenge due to privacy concerns.

As stated before, the mmW scanners deployed in airports used active radiation to acquire the mmW image. Due to the unavailability of active images for research purposes, this work has been focused on passive mmW images. Experiments conducted in this work show that to some extent passive mmW images can be used for person recognition. As discussed earlier, the quality of active images are better than passive images. Hence, our conjecture is that the proposed feature extraction method would also work well on active mmW images for person recognition. In addition, in Section 6.5 experiments conducted on the NIR images also show the significance of applying the proposed feature extraction and their fusion methods for person recognition. Hence, our approach is more generic and can be applied to recognize humans from images collected in different spectral bands.

An another important aspect to consider in future works on this topic is to carry out an analysis of the impact of the transmissivity of different types of clothing materials on mmW person recognition. Due to the limitations of the mmW TNO database (there was not clothing variation among the 4 different scenarios) only a few insights were gained in that regard, mainly 1) the more discrimination capabilities of the upper body part with respect to the lower body part (due to the thicker clothe material worn), and 2) the minor effect that facial accesories or balaclavas have only a minor impact on the mmW waves [23]. In [37], an experimental study was carried out to analyze the transmissivity of different types of clothing materials for passive millimeter wave images more specifically for concealed weapon detection. In that study, they considered materials such as cotton, wool and plastic with different thickness, and they concluded that mmW can readily penetrate different kinds of clothing material, being plastic material the most effective attenuating the capability of mmW, but not so much for the others considered even with many layers.

In any case, even with practical limitations, the present work has shown the feasibility of using mmW body texture for Person Recognition, and a quantitative analysis of the key factors affecting its performance, in this way, showing that current mmW security deployments may be augmented with biometrics functionalities.

## ACKNOWLEDGMENTS

This work was funded by project CogniMetrics (TEC2015-70627-R) from MINECO/FEDER and the SPATEK network (TEC2015-68766-REDC). E. Gonzalez- Sosa is supported by a PhD scholarship from Universidad Autonoma de Madrid. Author Fernando Alonso-Fernandez thanks the Swedish Research Council, the CAISR program and the SIDUS-AIR project of the Swedish Knowledge Foundation. Vishal M. Patel was partially supported by US Office of Naval Research (ONR) Grant YIP N00014-16-1-3134. Authors wish to thank also TNO for providing access to the database. The conclusion goes here.

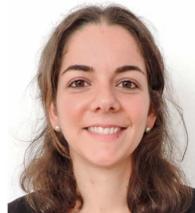

**Ester Gonzalez-Sosa** received her M.Sc in Electrical Engineering from Universidad de Las Palmas de Gran Canaria in 2012, her Bachelor in Computer Science in 2014 and her PhD from Universidad Autonoma de Madrid in 2017, the latter within the Biometrics and Data Pattern Analytics (BiDA-Lab) entitled "Face and Body Biometrics in the Wild: Advances in the Visible Spectrum and Beyond", reaching Cum Laude. She has carried out several research internships in worldwide leading groups in biometric recognition such as TNO, EURECOM, or Rutgers University. Since 2017, she joined the Distributed Reality Solutions team in Bell Labs Madrid. Her research interests include machine learning, computer vision, pattern recognition, deep learning to be applied in mixed realities or in biometric recognition. She has been the recipient of several awards such as the Uniteco Award from COIT, ICB-2018 Honorable Mention Award or EAB Research Award.

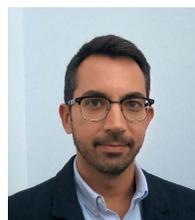

**Ruben Vera-Rodriguez** received the M.Sc. degree in telecommunications engineering from Universidad de Sevilla, Spain, in 2006, and the Ph.D. degree in electrical and electronic engineering from Swansea University, U.K., in 2010. Since 2010, he has been affiliated with the Biometric Recognition Group, Universidad Autonoma de Madrid, Spain, where he is currently an Associate Professor since 2018. His research interests include signal and image processing, pattern recognition, and biometrics, with emphasis on signature, face, gait verification and forensic applications of biometrics. Ruben has published over 85 research papers in high impact journal and international conferences. Dr. Vera-Rodriguez is actively involved in several National and European projects focused on biometrics. Ruben has been Program Chair for the IEEE 51st International Carnahan Conference on Security and Technology (ICCST) in 2017; and the 23rd Iberoamerican Congress on Pattern Recognition (CIARP 2018) in 2018.

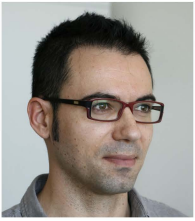

**Julian Fierrez** received the MSc and the PhD degrees in telecommunications engineering from Universidad Politecnica de Madrid, Spain, in 2001 and 2006, respectively. Since 2004 he has been affiliated with Universidad Autonoma de Madrid, where he is currently an Associate Professor since 2010. From 2007 to 2009 he was a visiting researcher at Michigan State University in USA under a Marie Curie fellowship. His research interests include general signal and image processing, pattern recognition, and biometrics, with emphasis on: handwriting, fingerprint, iris, face, security, privacy, HCI, behavior, forensics, and mobile forensic applications of biometrics. Prof. Fierrez has been actively involved in multiple EU projects focused on biometrics (e.g. TABULA RASA and BEAT), has attracted notable impact for his research, and is the recipient of a number of distinctions, including: EBF European Biometric Industry Award 2006, EURASIP Best PhD Award 2012, Miguel Catalan Award 2015 to the Best Researcher under 40 in the Community of Madrid in the general area of Science and Technology, and IAPR Young Biometrics Investigator Award 2017, given to a single researcher worldwide every two years under the age of 40, whose research work has had a major impact in biometrics. He is and Associate Editor for IEEE Trans. on Information Forensics and Security and IEEE Trans. on Image Processing.

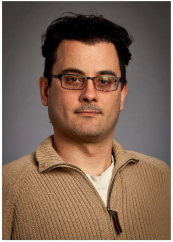

**Fernando Alonso-Fernandez** Fernando Alonso-Fernandez received the M.S. and Ph.D. degrees in telecommunications engineering from the Universidad Politecnica de Madrid, Spain, in 2003 and 2008, respectively. Since 2010, he has been with the Centre for Applied Intelligent Systems Research, Halmstad University, Sweden, first as the recipient of a Marie Curie IEF and a postdoctoral fellowship from the Swedish Research Council, and later as the recipient of a Project Research Grant for Junior Researchers of the Swedish Research Council. Since 2017, he is an Associate Professor at Halmstad University. He has been actively involved in in multiple EU (e.g. FP6 Biosecure NoE, COST IC1106) and National projects focused on biometrics and human-machine interaction. He also co-chaired ICB2016, the 9th IAPR International Conference on Biometrics. His research interests include signal and image processing, pattern recognition, and biometrics, with emphasis on facial cues and body biosignals. He has over 70 international contributions at refereed conferences and journals and has authored several book chapters.

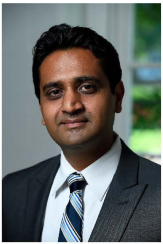

**Vishal Patel** Vishal M. Patel [SM'15] is an Assistant Professor in the Department of Electrical and Computer Engineering (ECE) at Johns Hopkins University. Prior to joining Hopkins, he was an A. Walter Tyson Assistant Professor in the Department of ECE at Rutgers University and a member of the research faculty at the University of Maryland Institute for Advanced Computer Studies (UMIACS). He completed his Ph.D. in Electrical Engineering from the University of Maryland, College Park, MD, in 2010. His current research interests include signal processing, computer vision, and pattern recognition with applications in biometrics and imaging. He has received a number of awards including the 2016 ONR Young Investigator Award, the 2016 Jimmy Lin Award for Invention, A. Walter Tyson Assistant Professorship Award, Best Paper Award at IEEE AVSS 2017, Best Paper Award at IEEE BTAS 2015, Honorable Mention Paper Award at IAPR ICB 2018, two Best Student Paper Awards at IAPR ICPR 2018, and Best Poster Awards at BTAS 2015 and 2016. He is an Associate Editor of the IEEE Signal Processing Magazine, IEEE Biometrics Compendium, and serves on the Information Forensics and Security Technical Committee of the IEEE Signal Processing Society. He is a member of Eta Kappa Nu, Pi Mu Epsilon, and Phi Beta Kappa.